\pdfoutput=1

\documentclass[11pt]{article}

\usepackage{EACL2023}

\usepackage{times}
\usepackage{latexsym}

\usepackage[T1]{fontenc}

\usepackage[utf8]{inputenc}

\usepackage{microtype}

\usepackage{inconsolata}

\usepackage{amsmath}
\usepackage{todonotes,booktabs,cleveref}
\usepackage{siunitx,subcaption,longtable,placeins}

\title{Towards a Computational Analysis of Suspense: \\ Detecting Dangerous Situations}

\author{Albin Zehe \\
University of Würzburg \\
\texttt{zehe@informatik.uni-wuerzburg.de} \\\And
Julian Schröter \\
University of Munich \\
\texttt{J.Schroeter@lmu.de} \\\AND
Andreas Hotho \\
University of Würzburg \\
\texttt{hotho@informatik.uni-wuerzburg.de}
\\}

\begin{document}
    \maketitle
    \begin{abstract}
        Suspense is an important tool in storytelling to keep readers engaged and wanting to read more.
        However, it has so far not been studied extensively in Computational Literary Studies.
        In this paper, we focus on one of the elements authors can use to build up suspense: dangerous situations.
        We introduce a corpus of texts annotated with dangerous situations, distinguishing between 7 types of danger.
        Additionally, we annotate parts of the text that describe fear experienced by a character, regardless of the actual presence of danger.
        We present experiments towards the automatic detection of these situations, finding that unsupervised baseline methods can provide valuable signals for the detection, but more complex methods are necessary for further analysis.
        Not unexpectedly, the description of danger and fear often relies heavily on the context, both local (e.g., situations where danger is only mentioned, but not actually present) and global (e.g., ``storm'' being used in a literal sense in an adventure novel, but metaphorically in a romance novel).
    \end{abstract}

    \section{Introduction}
    One of the main goals in recent computational literary studies research is to generate a computational model for the plot of a novel \cite{piper_narrative_2021, konle_modeling_2022}.
    While a full understanding is still out of range for current methods, we can study some important elements of plot.
    In this paper, we present work towards the computational analysis of suspense as one of these elements, specifically focusing on the detection of dangerous situations and situations where a character experiences fear.
    While it has often been emphasised that suspense is based on a lack of information and uncertainty, it is consensual that a protagonist being endangered is a necessary condition of suspense~\cite{hausenblas_spannung_2017}.
    Moreover, we hypothesise that the character's fear is, if not a necessary condition, a good indication of suspense if that character is indeed endangered in the fictional world.
    We maintain that the reader's fear for the protagonist's integrity is one of the most central criteria of suspense.
    For this aspect to be realised, dangerous situations are as crucial as the reader's uncertainty about the outcome.
    Moreover, dangerous situations mostly imply uncertainty.
    Hence, we try to provide an operationalisation of the detection of danger on the plot level and of characters fear as a basic step towards a more thorough model of suspense.
    In addition, we look at the aspect of character's fear in order to get a clearer understanding of the empirical correlation between both techniques and their relation to reading experiences of suspense based on expert annotation.

    \section{Related Work}
    \label{sec:related}
    Foundational theoretical research on the phenomenon of suspense has been conducted within the field of psychology and philosophy~\cite{levine_arousal_1967, tannenbaum_anatomy_1980, brewer_stories_1982, zillmann_logic_1991, bryant_responding_1991, vorderer_psychology_1996, wulff_spannungserleben_2002} and then unfolded in literary studies and linguistics~\cite{anz_spannung_2003, irsigler_spannungen_2008, ackermann_gespannte_2007, hausenblas_spannung_2017}.
    In the last decade, computational linguistics, particularly from the perspective of text generation, approached formalisation and models of suspense~\cite{cheong_suspenser_2015, oneill_dramatis_2014, algee-hewitt_machinery_nodate, doust_model_2017}.
    Although some of the latter works provide ambitious attempts to model uncertainty, they do not model the foundational dimension of dangerous situations on the plot level and fear on the level of fictional characters.

    \section{Detecting Dangerous Situations}
    \label{sec:task}
    In this paper, we present early experiments towards the detection of suspense.
    We focus on two tasks: detecting dangerous situations and detecting fear.

    \subsection{Annotation}
    Our texts were annotated with dangerous situations and fear descriptions.
    For fear, we only annotate situations where any character experiences fear, but no more specific information.
    For dangerous situations, we provide more fine-grained annotations regarding the specific type of danger.
    We define the types \texttt{Duel}, \texttt{Abduction}, \texttt{Natural}, \texttt{Supernatural}, \texttt{Ambush}, \texttt{Hitchcock}.
    For a more detailed description, see \cref{app:danger_types}.
    Previous work has shown that for corpora of highly stereotypical texts such as the German \textit{Heftroman} (dime novel) these types of plot aspects cover most of the relevant types of dangerous situations in several sub-genres~\cite{schroter_spannung_2023}.
    While the matter of generalisability for this set of danger sub-types for different domains requires further examination, our approach will allow us to validate how well each of the danger types can be operationalised respectively within this domain.

    \subsection{Classification Task}
    We define the task of detecting dangerous situations as a classification task:
    Given a unit of text, the task is to detect whether there is a dangerous situation present in the given unit, leading to a binary classification with the labels \texttt{Danger} and \texttt{NoDanger}.\footnote{For this early stage of our research, we do not distinguish between different types of danger.}
    Similarly, we define the task of detecting fear as a classification task with the labels \texttt{Fear} and \texttt{NoFear}.

    \section{Data}
    \label{sec:data}

    \paragraph{Raw Dataset}
    From a larger corpus of 19th century novellas and 20th century \textit{Heftromane} (dime novels) in German, student assistants and the authors annotated six texts, with three of these texts being annotated twice by different annotators each.

    \paragraph{Preprocessing}
    The texts were split into topically coherent segments, which we refer to as \emph{paragraphs} in the following, using the TextTiling algorithm~\cite{Hearst_CL97}, as implemented in the nltk Python library~\cite{Bird:2006:NNL:1225403.1225421}.
    We used the default hyperparameters for the algorithm, since we are only interested in a rough segmentation of the texts.

    \paragraph{Annotation}
    On a paragraph level, the different types of dangerous situations and the presence of character's fear were annotated for a total of 391 paragraphs.
    As the project of suspense analysis is at the beginning, we start with this small set of annotated texts.

    \paragraph{Resulting Dataset}
    \Cref{tab:situation_stats} in \cref{app:dataset_stats} shows the number of annotated paragraphs for each type of dangerous situation and fear description.
    We computed Cohen's kappa~\cite{cohen1960coefficient} for the inter-annotator agreement of the different annotators.
    The average kappa over all texts is \num{0.55} for the detection of dangerous situations, corresponding to a moderate agreement~\cite{landis1977measurement}, and \num{0.64} for the detection of fear descriptions, corresponding to a substantial agreement.
    If we alleviate the requirement to annotate the \emph{same type} of dangerous situation to only annotating whether there is \emph{any} dangerous situation present, the average kappa for dangerous situations increases to \num{0.61}, corresponding to a substantial agreement.

    \section{Methodology}
    \label{sec:methodology}
    We compare unsupervised baseline methods for the detection of dangerous situations and fear descriptions based on manually crafted word lists, which are automatically expanded by different methods.

    \subsection{Word Lists}
    Initially, we create lists of words we expect to be related to the different types of dangerous situations or to a general description of fear.
    These lists are then expanded by different techniques, based either on word embeddings or on the Knowledge Graph ConceptNet~\cite{speer2016conceptnet}, as described below.
    \Cref{tab:wordlists} shows the number of words contained in the base lists and the expanded lists.
    The ``Danger'' list is created by merging the lists for all subtypes of dangerous situations.
    Statistics for the subtype lists are given in \cref{app:word_lists} in \Cref{tab:word_lists_detail}.

    \paragraph{Word Embedding-based Expansion}
    To expand the word lists based on word embeddings, we use spacy's \texttt{de\_core\_news\_lg} model.
    For each word in the base list, we retrieved the 50 most similar words according to the word2vec model.
    We extended the word lists with the lemmas of these most similar words.

    \paragraph{ConceptNet-based Expansion}
    To expand the word lists based on ConceptNet, we used the ConceptNet API to retrieve all words related to the base words.
    We then selected all German words $B$ that are related to a base word $A$ by the relation $(A, \texttt{Synonym}, B)$ or $(B, \texttt{IsA}, A)$ that is, selecting words that are either equivalent to the base word or more specific.
    We discard ConceptNet nodes that contain spaces, since we are currently only matching single words.

    \begin{table}
        \centering
        \begin{tabular}{lrrr}
\toprule
{} &  Base &  Embeddings &  ConceptNet \\
Type       &       &             &             \\
\midrule
Fear       &    49 &          80 &         157 \\
Danger &   153 &         355 &         596 \\
\bottomrule
\end{tabular}

        \caption{Counts for the word lists for danger and fear. Base lists are manually created, Embedding and ConceptNet lists are expanded by the respective methods.}
        \label{tab:wordlists}
    \end{table}

    \subsection{Detecting Dangerous Situations}
    \label{sec:dangerous_situations}
    We use the word lists to detect dangerous situations and fear descriptions in the texts.
    For each unit $u$ (in our case, paragraphs) in the text, after lemmatisation, we count the number of words $w$ from each list $l_\text{type}$ that occur in the paragraph:
    \begin{equation}
        \text{danger}_\text{type}(u) = \|\{w \in u \mid w \in l_\text{type}\}\|.
    \end{equation}
    If $\text{danger}_\text{type}(u)$ for a unit $u$ is greater than the average over all units, we detect a dangerous situation of that type in the unit.
    As mentioned above, we currently only use the two types ``Fear Description'' and ``Any Dangerous Situation''.

    \section{Results}\label{sec:results}

    \begin{table*}[t]
        \centering
        \begin{tabular}{l|rrr|rrr}\toprule
 & \multicolumn{3}{c}{Dangerous Situation} & \multicolumn{3}{|c}{Fear Description} \\\cmidrule(lr){2-4} \cmidrule(lr){5-7}Word List & Precision & Recall & F1 & Precision & Recall & F1 \\\midrule
Base & 40.8 & 55.7 & 47.1 & 44.5 & 52.7 & 48.3 \\
Embeddings & 38.5 & 64.8 & 48.3 & 46.3 & 66.7 & 54.6 \\
ConceptNet & 30.3 & 83.0 & 44.4 & 35.4 & 55.9 & 43.3 \\
\bottomrule\end{tabular}
        \caption{F1-scores, Precision and Recall for the detection of dangerous situations (DS) and fear descriptions (FD).}
        \label{tab:results}
    \end{table*}

    \Cref{tab:results} shows the F1-scores, Precision and Recall for the detection of dangerous situations and fear descriptions with the different word lists.
    We find that, while the results are overall not yet good enough for further analysis, the word lists do provide signals towards the detection of dangerous situations and fear descriptions.
    For dangerous situations, as expected, expanding the base lists with either word embeddings or ConceptNet improves the recall, but decreases the precision.
    While the embedding-expanded list leads to a slightly higher F1-score than the base list, the ConceptNet expansion, even though limited to only synonyms and more specific words, seems to add too many unsuitable words, leading to a rather high recall, but very low precision.
    For fear descriptions, the results are similar, with the embedding-expanded list again reaching the highest F1-score.

    \paragraph{Quantitative Error Analysis}
    We performed two types of error analysis:
    First, we automatically extracted words that were responsible for false positives in the best-performing setting, that is, using embedding-expanded word lists.
    \Cref{tab:false_positives} shows the top 10 words that were responsible for false positives for the detection of dangerous situations.
    On the other hand, \Cref{tab:true_positives} shows the top 10 words that were responsible for true positives for the detection of dangerous situations.
    Inspecting some of the false positives, we find that the words are often used to describe the aftermath of a dangerous situation, e.g., blood (``Blut'') still dripping from a wound at a crime scene, or the victim (``Opfer'') being found by the police.
    The word murder (``Mord'') causes exclusively false positives in our dataset, likely because it is usually not mentioned in the process.
    On the other hand, words like knife (``Messer'') or blade (``Klinge'') seem to be very good indicators of dangerous situations.
    \Cref{tab:word_ratio,tab:word_ratio_cont} show the values for all words, while \Cref{tab:word_ratio_fear} contains the same information for the detection of fear descriptions.

    \begin{table*}
        \begin{subfigure}{0.49\textwidth}
            \centering
            \begin{tabular}{lrr}
\toprule
{} &  False Positives &  TP Ratio \\
word      &                  &           \\
\midrule
Blut      &               12 &      0.60 \\
schlagen  &               12 &      0.69 \\
Mord      &                9 &      0.00 \\
Opfer     &                7 &      0.75 \\
Wunde     &                5 &      0.55 \\
Regen     &                5 &      0.29 \\
töten     &                5 &      0.29 \\
wild      &                4 &      0.67 \\
gegenüber &                4 &      0.50 \\
Wind      &                4 &      0.85 \\
\bottomrule
\end{tabular}

            \caption{Words causing false positives}
            \label{tab:false_positives}
        \end{subfigure}
        \begin{subfigure}{0.49\textwidth}
            \centering
            \begin{tabular}{lrr}
\toprule
{} &  True Positives &  TP Ratio \\
word     &                 &           \\
\midrule
Messer   &              30 &      0.88 \\
schlagen &              27 &      0.69 \\
Wind     &              22 &      0.85 \\
Opfer    &              21 &      0.75 \\
Blut     &              18 &      0.60 \\
Klinge   &              18 &      0.90 \\
Sturm    &              12 &      0.92 \\
stoßen   &              10 &      0.77 \\
Feuer    &              10 &      0.83 \\
Fregatte &              10 &      0.91 \\
\bottomrule
\end{tabular}

            \caption{Words causing true positives}
            \label{tab:true_positives}
        \end{subfigure}
        \caption{Words causing most false positives (left) and true positives (right) for the detection of dangerous situations.}
    \end{table*}

    \paragraph{Qualitative Error Analysis}

    In order to enrich the quantitative error analysis, we conducted hermeneutic case studies based on an automatised recognition of false negatives and false positives including the words that lead to false positive assignments.
    Here, we could casually see strong domain specific dependencies for the reliability of the different words or even word groups.
    For instance, while some of the words that are pretty reliably indicative of dangerous situations on sea (such as ``Sturm'', ``Gewitter'', ``Orkan'' from the storm-list), these words occur also with high frequency in some love novels in a metaphorical use to express the intensity of love and passion.
    Similar issues of high versus low precision depending on the domain can be observed for verbs such as ``schlagen''.

    \paragraph{The Way Forward}
    Our results point out multiple directions for future work.
    We found that the word lists are - expectedly - not able to distinguish between different uses of words that are related to dangerous situations.
    This can either be because they are only mentioned after a dangerous situation has already occurred, or because they are used in a metaphorical sense.
    For both cases, training classifiers based on language models can help, since they should be able to capture the context of the words and thereby distinguish between different uses.
    We have already performed first experiments in this direction, however, our set of annotated texts is currently too small to be reasonably split into a training and evaluation set.
    We are currently working on expanding our annotated corpus.
    In addition, we also plan to make use of corpora annotated with emotion information, since emotions are strongly related to both descriptions of danger and fear -- indeed, fear is often considered a basic emotion~\cite{ekman1992basic,plutchik1980emotions}.
    However, the only dataset annotated with these basic emotions in German literature we are aware of~\cite{zehe2017sentiment} only contains a small number of samples annotated with ``fear''.
    We will also explore using English datasets annotated with these emotions, as for example~\citet{Kim2019}.

    \section{Discussion and Conclusion}\label{sec:discussion-and-conclusion}
    From a general point of view, our paper contributes mainly to one dimension of the structure of a plot based understanding of suspense and prepares the ground for exploring a second – and more reader response based – dimension.
    These dimensions can be described as (a) the operationalisability of dangerous situations on the plot level by means of vocabulary based methods, and (b) the extent to which the psychological experience of suspense can be approximated by danger and the characters' fear.
    With regard to the first dimension (a), we were able to show that vocabulary based methods are capable of capturing a significant signal and can be regarded as baselines methods for future research.
    For future work, our results point the way to improving the methods.
    We also saw that different wordlists differ in their performance relative to the semantic domains they are applied in.
    Regarding the second dimension (b) and on a level of casual analysis, we found that some texts combine a high degree of danger with a low level of the protagonist's fear.
    A systematic correlation study of danger and fear with reader annotations of suspense experience can help to identify and better understand different types of suspense such as the Hitchcock-type, wherein the characters are not in fear because they don't realise being endangered, or the action-hero suspense, wherein the hero won't be in fear when being endangered due to his personal character, or special types of psychological suspense narration, where the – neurotic – characters are in constant fear without actually being endangered.
    An important step for future work is to provide a structural combination of the levels of emotion (characters' fear), plot (dangerous situation) and uncertainty on the level of reader response.

    \section*{Acknowledgements}
    Parts of this contribution resulted from the project “A Mixed Methods Design for Computational Genre Stylistics and Unstructured Genres. Towards a Functional History of 19th Century German Novellas” (project number 449668519), which was funded by the Deutsche Forschungsgemeinschaft (DFG) as a Walter-Benjamin Fellowship.

    \bibliography{anthology,custom}
    \bibliographystyle{acl_natbib}

    \appendix

    \section{Appendix}

    \subsection{Types of Danger}\label{app:danger_types}
    The different types of dangerous situations, which were operationalised based on word lists, start with a qualitative understanding of prototypical situations. \texttt{Duel} is regarded as a symmetrical fight between two antagonists, \texttt{Abduction} as the asymmetrical event of a kidnapping, \texttt{Natural} as a storm, fire, or another natural disaster threatening the protagonist's life, \texttt{Supernatural} as an indeterminate threat mostly coming from demonic power, \texttt{Ambush} as an asymmetric attack with a clear distinction between offender and victim, and \texttt{Hitchcock} as the type of suspense with a knowledge divide between actual audience and fictional character. Hence, this type of suspense is located on a different theoretical level of reader response as it is not defined by a specific plot type but rather by the aspect that the protagonist is unaware of being threatened while the audience is in the knowledge of the danger. We wanted to cover this latter type of suspense because it is central to the understanding of suspense in literary and cultural studies. The fact that it hasn't even been seen in the annotation set (see table 6) entails, however, that this type of suspense had to be ignored in our analyses.

    \subsection{Dataset Statistics}
    \label{app:dataset_stats}
    \FloatBarrier
    \begin{table}[h!]
        \begin{tabular}{lr}
\toprule
{} &  Count \\
\midrule
DangerousSituationNatural      &     36 \\
DangerousSituationOther        &     14 \\
DangerousSituationDuel         &     10 \\
DangerousSituationSupernatural &     12 \\
DangerousSituationAmbush       &     16 \\
FearDescription                &    104 \\
\bottomrule
\end{tabular}

        \caption{Counts of the different types of dangerous situations and fear descriptions annotated in our dataset.}
        \label{tab:situation_stats}
    \end{table}
    \FloatBarrier
    \vfill\eject

    \subsection{Word Lists}
    \label{app:word_lists}
    \FloatBarrier
    \begin{table}[h!]
        \centering
        \begin{tabular}{lrrr}
\toprule
{} &  Base &  Embeddings &  ConceptNet \\
Type       &       &             &             \\
\midrule
Fear       &    49 &          80 &         157 \\\midrule
Any Danger &   153 &         355 &         596 \\\midrule
Abduction  &    15 &          35 &          34 \\
Fire       &    24 &          58 &          76 \\
Violence   &    27 &          83 &         107 \\
War        &    35 &          76 &         179 \\
Storm      &    34 &          69 &         120 \\
Duel       &    25 &          51 &         112 \\
\bottomrule
\end{tabular}

        \caption{Counts for the word lists for the detection of dangerous situations. Base lists are manually created, Embeddings lists are created by expanding the base lists with words similar to the base words, and ConceptNet lists are created by expanding the base lists with words related to the base words in ConceptNet.}
        \label{tab:word_lists_detail}
    \end{table}
    \FloatBarrier
    \vfill\eject

    \subsection{Error Analysis}
    \label{app:error_analysis}
    \FloatBarrier
    \begin{table}[h!]
        \centering
        \begin{tabular}{lrrr}
\toprule
{} &  TP &  FP &  TP Ratio \\
word            &     &     &           \\
\midrule
würgen          &   2 &   0 &      1.00 \\
donnern         &   2 &   0 &      1.00 \\
brennend        &   2 &   0 &      1.00 \\
Bö              &   3 &   0 &      1.00 \\
rammen          &   2 &   0 &      1.00 \\
röcheln         &   2 &   0 &      1.00 \\
zuschlagen      &   3 &   0 &      1.00 \\
zerren          &   4 &   0 &      1.00 \\
Hagel           &   2 &   0 &      1.00 \\
Kampf           &   6 &   0 &      1.00 \\
Donner          &   2 &   0 &      1.00 \\
schießen        &   8 &   0 &      1.00 \\
verschlingen    &   2 &   0 &      1.00 \\
Sieger          &   4 &   0 &      1.00 \\
explodieren     &   2 &   0 &      1.00 \\
erwürgen        &   2 &   0 &      1.00 \\
Sturmwind       &   2 &   0 &      1.00 \\
entfliehen      &   2 &   0 &      1.00 \\
entstellt       &   2 &   0 &      1.00 \\
gefangen        &   2 &   0 &      1.00 \\
Sturm           &  12 &   1 &      0.92 \\
Fregatte        &  10 &   1 &      0.91 \\
Klinge          &  18 &   2 &      0.90 \\
Messer          &  30 &   4 &      0.88 \\
Blitz           &   6 &   1 &      0.86 \\
fliehen         &   6 &   1 &      0.86 \\
Wind            &  22 &   4 &      0.85 \\
Feuer           &  10 &   2 &      0.83 \\
packen          &   4 &   1 &      0.80 \\
zünden          &   8 &   2 &      0.80 \\
stoßen          &  10 &   3 &      0.77 \\
Waffe           &   6 &   2 &      0.75 \\
Opfer           &  21 &   7 &      0.75 \\
schlagen        &  27 &  12 &      0.69 \\
umbringen       &   2 &   1 &      0.67 \\
wild            &   8 &   4 &      0.67 \\
gegenüberstehen &   2 &   1 &      0.67 \\
brennen         &   6 &   3 &      0.67 \\
schleudern      &   6 &   3 &      0.67 \\
Explosion       &   2 &   1 &      0.67 \\
\bottomrule
\end{tabular}

        \caption{True Positive, False Positive and True Positive Ratio for Dangerous Situations caused all words that were used in the corpus}
        \label{tab:word_ratio}
    \end{table}

    \begin{table}[t!]
        \centering
        \begin{tabular}{lrrr}
\toprule
{} &  TP &  FP &  TP Ratio \\
word          &     &     &           \\
\midrule
Blut          &  18 &  12 &      0.60 \\
Wunde         &   6 &   5 &      0.55 \\
gegenüber     &   4 &   4 &      0.50 \\
Welle         &   2 &   3 &      0.40 \\
Regen         &   2 &   5 &      0.29 \\
töten         &   2 &   5 &      0.29 \\
blutend       &   0 &   1 &      0.00 \\
Stellung      &   0 &   1 &      0.00 \\
Streit        &   0 &   1 &      0.00 \\
beißen        &   0 &   1 &      0.00 \\
graben        &   0 &   1 &      0.00 \\
blutig        &   0 &   1 &      0.00 \\
ermorden      &   0 &   2 &      0.00 \\
flammen       &   0 &   1 &      0.00 \\
Pistole       &   0 &   1 &      0.00 \\
kämpfen       &   0 &   1 &      0.00 \\
Prasseln      &   0 &   1 &      0.00 \\
Feind         &   0 &   1 &      0.00 \\
Mord          &   0 &   9 &      0.00 \\
Gewitter      &   0 &   3 &      0.00 \\
Gewalt        &   0 &   2 &      0.00 \\
löschen       &   0 &   1 &      0.00 \\
morden        &   0 &   2 &      0.00 \\
durchbohren   &   0 &   1 &      0.00 \\
vergewaltigen &   0 &   1 &      0.00 \\
regnen        &   0 &   3 &      0.00 \\
prasseln      &   0 &   1 &      0.00 \\
sperren       &   0 &   1 &      0.00 \\
überfallen    &   0 &   1 &      0.00 \\
\bottomrule
\end{tabular}

        \caption{Continuation of \Cref{tab:word_ratio}}
        \label{tab:word_ratio_cont}
    \end{table}

    \vfill\null\eject

    \begin{table}[t!]
        \centering
        \begin{tabular}{lrrr}
\toprule
{} &  TP &  FP &  TP Ratio \\
word         &     &     &           \\
\midrule
bedrohen     &   5 &   0 &      1.00 \\
widerlich    &   1 &   0 &      1.00 \\
ängstlich    &   2 &   0 &      1.00 \\
nervös       &   1 &   0 &      1.00 \\
höllisch     &   1 &   0 &      1.00 \\
Bedrohung    &   1 &   0 &      1.00 \\
verwirren    &   1 &   0 &      1.00 \\
erbärmlich   &   1 &   0 &      1.00 \\
Angstschweiß &   1 &   0 &      1.00 \\
Panik        &   1 &   0 &      1.00 \\
zittern      &   9 &   1 &      0.90 \\
Furcht       &   6 &   1 &      0.86 \\
Angst        &  20 &   4 &      0.83 \\
schrecken    &   3 &   1 &      0.75 \\
Zittern      &   5 &   2 &      0.71 \\
ohnmächtig   &   2 &   1 &      0.67 \\
Schrecken    &   3 &   2 &      0.60 \\
erschrecken  &   3 &   3 &      0.50 \\
fürchten     &   2 &   2 &      0.50 \\
schrecklich  &  15 &  15 &      0.50 \\
Gänsehaut    &   1 &   1 &      0.50 \\
Schreck      &   2 &   2 &      0.50 \\
furchtbar    &   3 &   5 &      0.38 \\
gefährlich   &   2 &   4 &      0.33 \\
leiden       &   2 &   5 &      0.29 \\
Gefahr       &   3 &   9 &      0.25 \\
schlimm      &   4 &  13 &      0.24 \\
lähmen       &   0 &   1 &      0.00 \\
hoffnungslos &   0 &   1 &      0.00 \\
grässlich    &   0 &   1 &      0.00 \\
schauerlich  &   0 &   1 &      0.00 \\
schocken     &   0 &   1 &      0.00 \\
fürchterlich &   0 &   3 &      0.00 \\
wehrlos      &   0 &   1 &      0.00 \\
erschreckt   &   0 &   1 &      0.00 \\
unheimlich   &   0 &   4 &      0.00 \\
besorgt      &   0 &   1 &      0.00 \\
bedrohlich   &   0 &   1 &      0.00 \\
Leiden       &   0 &   1 &      0.00 \\
schwitzen    &   0 &   1 &      0.00 \\
überwältigen &   0 &   1 &      0.00 \\
\bottomrule
\end{tabular}

        \caption{True Positive, False Positive and True Positive Ratio for Fear Descriptions caused all words that were used in the corpus}
        \label{tab:word_ratio_fear}
    \end{table}
    \FloatBarrier

\end{document}